\documentclass{article}
\usepackage{spconf,amsmath,graphicx}
\usepackage{multirow}
\usepackage{amssymb}
\usepackage{xcolor}  
\usepackage{cite}

\title{
End-to-end spoken language understanding using joint CTC loss and self-supervised, pretrained acoustic encoders
}
\name{
Jixuan Wang \qquad
Martin Radfar \qquad
Kai Wei \qquad
Clement Chung \qquad
}

\address{Amazon Alexa AI}
\begin{document}
%
\maketitle
\begin{abstract}
It is challenging to extract semantic meanings directly from audio signals in spoken language understanding (SLU), due to the lack of textual information. Popular end-to-end (E2E) SLU models utilize sequence-to-sequence automatic speech recognition (ASR) models to extract textual embeddings as input to infer semantics, which, however, require computationally expensive auto-regressive decoding. In this work, we leverage self-supervised acoustic encoders fine-tuned with Connectionist Temporal Classification (CTC) to extract textual embeddings and use  joint CTC and SLU losses for utterance-level SLU tasks.  Experiments show that our model  achieves 4\% absolute improvement over the the state-of-the-art (SOTA) dialogue act classification model on the DSTC2 dataset and 1.3\% absolute improvement over the SOTA SLU model on the SLURP dataset.




\end{abstract}
\section{Introduction}
\label{sec:intro}

Spoken language understanding (SLU) models infer semantics from spoken utterances~\cite{wang2005spoken, TurDeMori2011,Bhargava2013EasyCI}.
Common SLU tasks include intent classification, slot filling and dialogue act classification.
Traditionally, SLU models consist of an ASR model that transcribes audio signals into text and a natural language understanding (NLU) model that extracts semantics from the text~\cite{larson2012spoken}.
Since the ASR and NLU models are optimized independently, the errors from the ASR component will be propagated to the NLU component. For example, if ``turn on TV" is recognized as ``turn off TV" by ASR, it will be challenging for NLU to predict the right intent.

Recently, there is growing interest in building E2E SLU models where the acoustic and textual models are jointly optimized, leading to more robust SLU models~\cite{lugosch2019speech, serdyuk2018towards, agrawal2022tie, kim2021st, haghani2018audio, kuo2020end, radfar2020end, radfar2021fans, rao2020speech, saxon2021end, seo2022integration, raju2022joint}.
Early works~\cite{lugosch2019speech, serdyuk2018towards} learn an utterance-level semantic representation directly from audio signals without performing speech recognition. This is challenging because semantic information may be lost without the textual information. 
To guide the learning of semantic representations from audio signals, multi-modal losses are used to tie the utterance-level embeddings from a pretrained BERT model and an acoustic encoder~\cite{agrawal2022tie}. Token-level cross-modal alignment and joint space learning is studied in~\cite{kim2021st}. 


The limitation of the above approaches is that they cannot be used for sequence labeling tasks, like slot filling. To address this issue, another stream of works build unified models, which can be trained end-to-end and used for both intent classification and slot filling. One way to achieve E2E training is to re-frame SLU as a sequence-to-sequence task, where semantic labels are treated as another sequence of output labels besides the transcript~\cite{haghani2018audio, kuo2020end, radfar2020end, radfar2021fans}. 
Another way is to unify ASR and NLU models and train them together via differentiable neural interfaces~\cite{rao2020speech, saxon2021end, seo2022integration, raju2022joint}. 
One commonly used neural interface is to feed the token level hidden representations from ASR as input to the NLU model~\cite{rao2020speech, saxon2021end, seo2022integration, raju2022joint}. \cite{saxon2021end, seo2022integration}  utilizes pretrained ASR and NLU models with shared vocabulary.
Different neural interfaces are compared and a novel interface for RNN-Transducer (RNN-T) based ASR model is proposed in~\cite{raju2022joint}.
However, to produce token-level representations at inference time, those approaches need auto-regressive decoding, which is computationally expensive. 

In ASR models trained with the CTC loss, labels are predicted at the audio frame level in parallel,
which are more efficient than those requiring auto-regressive decoding~\cite{graves2012sequence,chan2016listen}. 
In this work, we use the output of a CTC-based ASR model as input to infer semantics and joint CTC and SLU losses to train the model end-to-end. 
We show that our approach outperforms both the approaches that infer semanctics directly from audio without ASR supervision and the approaches that rely on auto-regressive ASR models. 
Compared with the approach proposed in~\cite{chen2018spoken}, our work demonstrates the effectiveness of using acoustic encoders pretrained with self-supervised tasks for E2E SLU, and highlights the importance of joint training with both CTC and SLU losses and the use of logits instead of probabilities as input to extract utterance embeddings.
We show that this simple approach achieves SOTA results on three public datasets of three different tasks. Notably, our model outperforms the best reported intent classification accuracy on the SLURP dataset by 1.3\%.

\section{E2E SLU with CTC}
\begin{figure}[tb]
\begin{minipage}[b]{1.0\linewidth}
  \centering
  \centerline{\includegraphics[width=8cm]{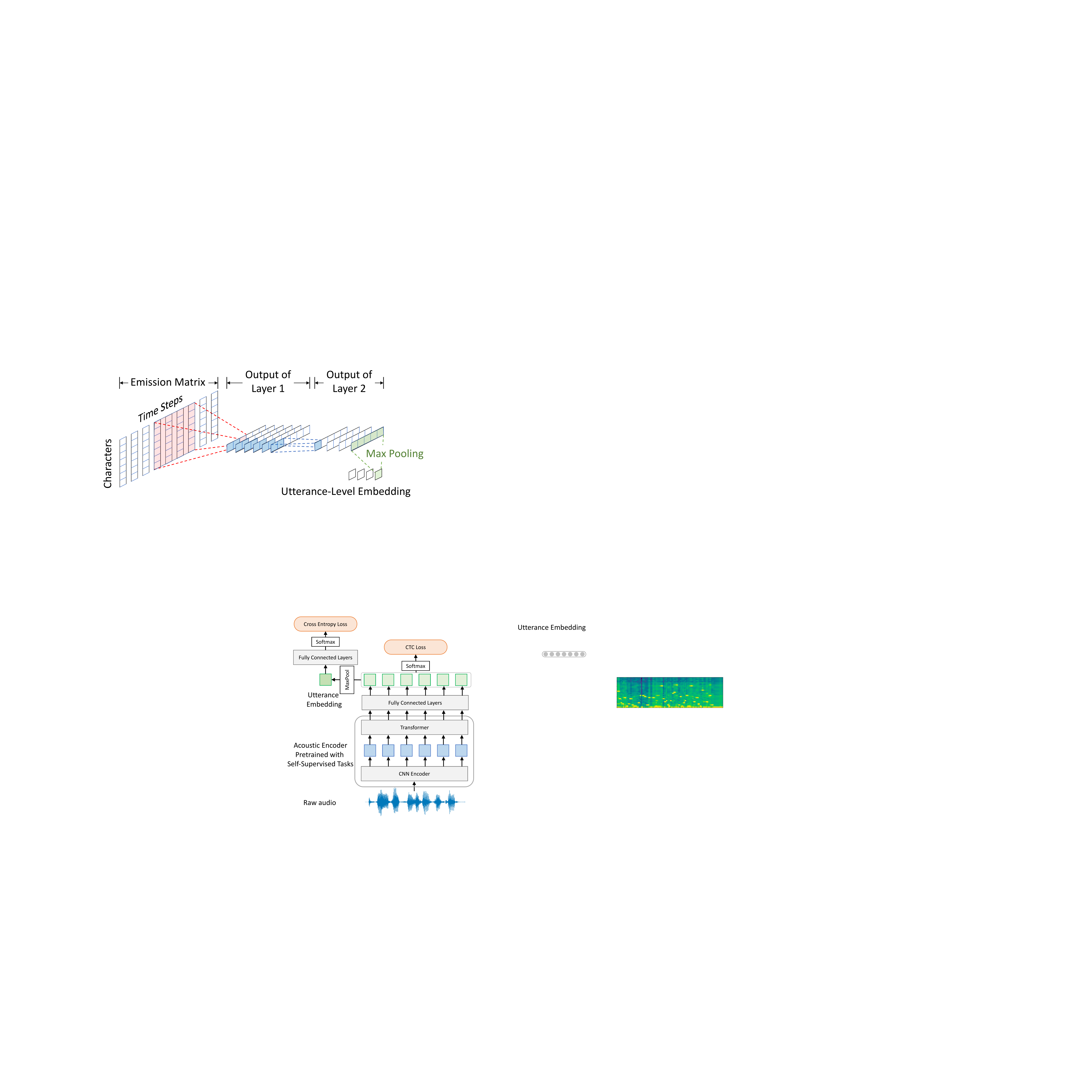}}
\end{minipage}
%
\caption{
The block diagram of the proposed approach.
}
\label{fig:main}
\end{figure}

Our model mainly consists of two parts as shown in Figure~\ref{fig:main}: an ASR model and an utterance encoder. The ASR model is based on an acoustic encoder pretrained by self-supervised tasks and fine-tuned using CTC.
In this work, we leverage the pretrained Wav2Vec2.0~\cite{baevski2020wav2vec} and HuBERT~\cite{hsu2021hubert} models.
The output sequence of the ASR model is maxpooled and fed into the utterance encoder, which is based on fully connected layers. 
The whole model can be trained end-to-end.

Let’s denote the audio sequence of an utterance as $X = (x_1, x_2, \dots, x_{T}) $ and the corresponding transcript as $W = (w_1, w_2, \dots, w_U)$, where $T$ and $U$ are the sequence length of the acoustic features and transcript, respectively. Each utterance is annotated with an utterance-level label $y$, such as intent or dialogue act. The training data is denoted by 
$D^\text{tr} = \{(X_i, W_i, y_i)\}_{i=1}^{|D^\text{tr}|}$.
To train an ASR model on $D^\text{tr}$, we would like to maximize the expected conditional probability $\mathbb{E}_{(X,W)\sim p^{tr}} p(W|X)$ with respect to the training data distribution $p^{tr}$, which is approximated by $\sum_{i=1}^{|D^\text{tr}|} p(W_i|X_i)$.

CTC 
 does not assume that the alignment between input and output sequences is given but considers all possible alignments when calculating the training loss.
For each pair of $(X, W)$, we calculate $p(W|X)$ as the sum of the probability of all valid alignments $ \mathcal{A}_{X, W}$:
\begin{equation}
    p(W|X) = \sum_{A \in \mathcal{A}_{X, W}} p(A|X) 
\end{equation}
where $A = (a_1, a_2, \dots, a_T)$ and each $a_i$ can take any value from the set of all possible output tokens and a special token $\epsilon$, which refers to a blank symbol.
By removing repeating tokens and $\epsilon$ from an alignment $A$, we can recover the output sequence $W$, then $A$ is regarded as a valid alignment. 

To calculate $p(A|X)$,  we use an acoustic encoder to extract a sequence of frame-level hidden representation $H = (h_1, \dots, h_{T'})$. Note $T'$ will be smaller than $T$ if the acoustic encoder contains subsampling layers, but we assume $T'=T$ for notation simplicity. 
Frame level prediction is given by: 
\begin{equation}
    p(a_i|X) = softmax(Wh_i + b), i = \{1, 2, \dots, T\}.
\end{equation}
where $W$ and $b$ are parameters of a linear classifier.

The conditional probability $ p(W|X)$ can be efficiently calculated through dynamic programming and used as training objective to optimize the ASR model. The ASR objective is defined by:
\begin{equation}
\mathcal{L}^{\text{CTC}} = \frac{1}{|D^\text{tr}|} \sum_{(W, X) \in D^\text{tr}} - \log p(W|X),
\end{equation}


To predict utterance level labels for tasks like intent classification and dialogue act classification,
we use the frame-level logits $(W h_1 + b, \dots, W h_T+b)$ as input to the utterance encoder.
We also try using the hidden representations $H$ as input to the utterance encoder.
We observe the two approaches perform similarly in the experiments.

Let's denote the input to the utterance encoder as $H^u = (h^u_1, \dots, h^u_T)$.
Our utterance encoder contains a maxpooling layer and two fully connected layers:
\begin{equation} \label{eq1}
\begin{split}
    h^{pool} &= maxpool(h_1^u, \dots, h_{T}^u) \\
    h^{utt} &= g(W^{(2)} g(W^{(1)} h^{pool} + b^{(1)}) + b^{(2)}) \\
\end{split}
\end{equation}
where $W^{(1)}, b^{(1)}$ and $W^{(2)}, b^{(2)}$ are the parameters of the two fully connected layers, respectively, and $g$ is an activation function. For classification, we use a linear classifier  and the cross entropy loss for training: 

\begin{equation} \label{eq2}
\begin{split}
p(y|X) &= softmax(W^u h^{utt} + b^u) \\
\mathcal{L}^\text{SLU} &= \frac{1}{|D^\text{tr}|} \sum_{(y, X) \in D^\text{tr}} -\log p(y|X).
\end{split}
\end{equation}

\noindent We train the whole model in a multi-task fashion using a linear combination of $\mathcal{L}^\text{CTC}$ and $\mathcal{L}^\text{SLU}$:
\begin{equation} \label{eq4}
\begin{split}
    \mathcal{L} = \alpha^\text{CTC} \mathcal{L}^\text{CTC} +  \alpha^\text{SLU} \mathcal{L}^\text{SLU}
\end{split}
\end{equation}
where we treat scalars $\alpha^\text{CTC}$ and $\alpha^\text{SLU}$ as hyperparameters.

Good ASR performance is the key to understanding the semantics, so we first fine-tune the ASR model using in-domain data during training, by setting $\alpha^\text{SLU} = 0, \alpha^\text{CTC}=1$. After the ASR model stops improving on the validation set, we set $\alpha^\text{SLU}>0, \alpha^\text{CTC}>0$, tune their values based on validation performance and train the whole model end-to-end. 

\section{Experiments}
We conduct experiments on dialogue act classification, keyword spotting and intent classification tasks, where the inputs are audio signals and the outputs are utterance-level labels.

\subsection{Datasets}
We use three public datasets listed in Table~\ref{datasets}, showing the number of utterances used for training, validation and testing for each dataset.

\noindent\textbf{Dialogue act classification:} We use the DSTC2 dataset~\cite{henderson2014second} for this task, where each utterance is annotated with dialogue acts containing action type, slot name and slot value. Similar to~\cite{wei2022neural}, we focus on utterance level classification without considering the adjacent utterances in the same dialogue and combine multiple labels appearing in the same utterance into a single label. 

\noindent\textbf{Keyword spotting:}
We use the Google Speech Commands (GSC) dataset V2~\cite{warden2018speech}, which has 105K snippets in total. Each audio snippet is 1 second long and contains a single word. There are 35 unique words in total that are also used as labels. The goal is to predict the word given an audio snippet. 

\noindent\textbf{Intent classification:}
For this task, we use the SLURP dataset~\cite{bastianelli2020slurp}.
SLURP is a challenging dataset with linguistically more diverse utterances and a bigger label space across 18 domains. Each utterance is annotated with a \textit{scenario} and \textit{action} label. Following previous works~\cite{bastianelli2020slurp, arora2022espnet}, we define \textit{intent} as the combination of \textit{scenario} and \textit{action}. In total, there are 69 unique intents in the training set. 

\subsection{Baselines}
For the \textit{dialogue act classification} task, we use the model proposed in~\cite{wei2022neural} as the baseline, which is the most recent work on this dataset. The authors introduce a neural prosody encoder for dialogue act classification that uses both prosodic features and raw audio signals as input. 
We follow the same procedure to prepare the DSTC2 dataset.

For the \textit{keyword spotting} task, we compare with a recent work called HTS-AT~\cite{chen2022hts}, which is transformer-based model with a hierarchical structure.

For \textit{intent classification}, we compare with the following SOTA models.
\textit{ESPNET-SLU:} The ESPnet-SLU toolkit has several SLU recipes for SLURP~\cite{arora2022espnet}. Their SLU model is a sequence-to-sequence model where the intent is decoded as one word. We include their results achieved by using pretrained HuBERT as feature extractors for SLURP. 
\textit{Seo et al., 2022:} This work used a Wav2Vec2.0-based ASR model~\cite{seo2022integration}. The output of the decoder is fed into a RoBERTa-based NLU model for intent classification and slot filling. Each component of this model is first pretrained separately and then fine-tuned jointly.
We compare with their results using both the original and synthetic training sets.

\begin{table}
\caption{\label{datasets}Dataset information.}
\begin{tabular}{ ccccc } 
 \hline
 \hline
 Task & Dataset & Train & Valid & Test \\ 
 \hline
 DAC & DSTC2 & 12,930 & 1,437 & 9,116 \\ 
 KWS & Speech Commands & 84,843 & 9,981 & 11,005 \\ 
 IC & SLURP-Synth & 119,881 &  8,690 &  13,078 \\ 
 \hline 
 \hline
\end{tabular}
\end{table}

\subsection{Implementation \& Hyperparameters}
\label{sec:hyper}
Our approach can benefit from any pretrained ASR models trained with the CTC loss.
We leverage the pretrained Wav2Vec2.0 [20] and HuBERT [21] acoustic encoders fine-tuned using CTC, as they were used by previous SOTA SLU models~\cite{arora2022espnet, seo2022integration}. Depending on the size of the datasets, we use a different sized pretrained ASR model.
We used the pretrained ASR models provided by TorchAudio~\cite{yang2021torchaudio}. For smaller datasets like DSTC2 and Speech Commands, we used the smallest models available in TorchAudio ``WAV2VEC2 ASR BASE 960H"~\cite{baevski2020wav2vec}. For more challenging SLURP dataset, we use ``HUBERT ASR LARGE"~\cite{hsu2021hubert}. Both models are pretrained on unlabeled audio data and fine-tuned on the LibriSpeech dataset~\cite{librispeech}.
For the input to the utterance encoder, we used maxpooling followed by two fully connected layers, each of which has 128 hidden units and uses the GELU activation function~\cite{hendrycks2016gaussian}.

When fine-tuning the ASR models, we stop training if the ASR loss has not been improved for 5 epochs. During joint training, we run training for 50 epochs and select the checkpoint with the best performance on the validation set. Since the tasks we study are all utterance-level classification tasks, we use accuracy as the metric for evaluation.
We tune learning rate, batch size and $\alpha^\text{CTC}$ based on validation performance. 
For DSTC2, we use learning rate of 0.00001 and batch size of 16 and $\alpha^\text{CTC}$ of 0.5.
For Speech Commands, we use learning rate of 0.00001 and batch size of 128 and $\alpha^\text{CTC}$ of 1.0.
For SLURP, we use learning rate of 0.00005, batch size of 128 and $\alpha^\text{CTC}$ of 0.5.
We use AdamW optimizer implemented in PyTorch with the default configurations.

\section{Results}
\subsection{Comparison with previous works}
The results on the three datasets are shown in Table~\ref{table:main}. 

\noindent
\textbf{Dialogue act classification:} 
Our model achieves 97.6\% and 97.5\% accuracy on DSTC2 using hidden representations and frame-level logits as input, respectively, which significantly outperforms the previous work~\cite{wei2022neural} that directly predicts dialogue acts from audio input by 4\%. 

\noindent
\textbf{Keyword Spotting:}
There is little improvement space on GSC V2 as previous work already achieves very good performance.
Although our approach is not designed for keyword spotting, it still matches the performance of HTS-AT~\cite{chen2022hts}, achieving 98.0\% accuracy. Note that all the utterances in GSC V2 only contain a single word, thus the textual information recovered by the ASR model in our approach might not be as helpful as on other datasets, like DSTC2 and SLURP, which contain linguistically more complex utterances.

\noindent
\textbf{Intent Classification:}
On the SLURP dataset, our approach achieves 88.1\% and 88.2\% accuracy using hidden representation and frame-level logits as input, respectively, resulting in 1.3\% absolute improvement over the SOTA result.

\begin{table}
\centering
\caption{\label{dstc2}Test results on DSTC2, Speech Commands and SLURP. *Numbers are obtained from the original papers.}
\begin{tabular}{ cp{0.27\textwidth}l } 
 \hline
 \hline
 Dataset                & Approach                  & Accuracy \\ 
 \hline
 \multirow{2}{*}{DSTC2} & Wei et al., 2022~\cite{wei2022neural} & 93.6* \\ 
                        & Ours (Wav2Vec2.0)         &  \\
                        & \quad + \textbf{Hidden as input}     & \textbf{97.6} \\ 
                        & \quad + Logits as input     & 97.5 \\ 
 \hline 
 \multirow{3}{*}{GSC}   & HTS-AT~\cite{chen2022hts}            & 98.0* \\
                        & Ours (Wav2Vec2.0)         &  \\
                        & \quad + Hidden as input   & 98.0 \\ 
                        & \quad + Logits as input   & 98.0 \\ 
 \hline 
\multirow{5}{*}{SLURP} & ESPnet-SLU~\cite{arora2022espnet}                 & 86.3*   \\ 
                        & Seo et al., 2022~\cite{seo2022integration}       & 86.9*  \\ 
                        & Ours (HuBERT)         &  \\
                        & \quad + Hidden as input     & 88.1  \\
                        & \quad \textbf{+ Logits as input}     & \textbf{88.2} \\ 
 \hline 
 \hline 
\end{tabular}
\label{table:main}
\end{table}

\subsection{Ablations \& Analysis}
We conduct ablation studies on the SLURP dataset to investigate: 1) whether our E2E models can perform better than cascade models; 2) whether CTC loss is useful for E2E SLU modeling; 3) whether fine-tuning the ASR model for SLU is necessary; 4) whether we need more powerful utterance encoders. Results are listed in Table~\ref{table:ablation}.

\textbf{HuBERT + BiLSTM/BERT NLU: }
We train BiLSTM and BERT-based NLU model on the one-best hypothesis generated by the fine-tuned HuBERT ASR model. The number of parameters in the BiLSTM and BERT NLU model is 24.7M and 110M, respectively. The textual encoder of our model only contains two fully connected layers (20K parameters), which is much smaller than the BiLSTM and BERT models and trained from scratch without pretraining on textual data as by BERT. Still, our approach outperforms the two cascade approaches: the test accuracy of the two approaches is 83.95\% and 87.44\%, respectively, while ours is 88.18\%.

\textbf{HuBERT + linear classifier w/o ASR loss:}
The only difference between this approach and our proposed approach is that this approach does not utilize the CTC loss for SLU but predicts SLU labels directly from audio signals.
The test accuracy of this approach is 84.81\%, which is lower than both our approach and the previous work based on the LAS ASR model~\cite{seo2022integration}. 
This result demonstrates the importance of the textual information recovered by the ASR loss for SLU.

\textbf{HuBERT (frozen) + utterance encoder:}
After fine-tuning the ASR model on the in-domain data, instead of further fine-tuning it with the SLU task, we freeze the ASR model and only train the utterance encoder. This approach only achieves 72.16\% accuracy. This shows that further fine-tuning the ASR model for SLU tasks in an E2E fashion is one of the keys to good performance.

\textbf{HuBERT with probabilities as input} Similar with previous work~\cite{chen2018spoken}, we experiment with using probabilities after the softmax layer as input to the utterance encoder. This approach performs worse than using logits or hidden representations as input, resulting in 87.0\% accuracy.

\textbf{HuBERT with CNN textual encoder: } We investigate whether using a more powerful CNN-based utterance encoder can lead to better performance. We apply four CNN layers with different kernel sizes on top of the logits and apply max pooling and fully-connected layers to extract the utterance-level embeddings. We observe that although the CNN encoder (1.9M parameters) is much larger than our utterance encoder (20K parameters), the accuracy using the CNN encoder is 88.05\% and slightly worse than our approach.

We further conduct analysis on ASR performance. 
The WER and CER of the ASR model before SLU training are 18.2\% and 8.5\%, respectively, while they decrease to 17.4\% and 7.8\%, respectively, after SLU training. This shows that the SLU loss does not conflict with the ASR loss but can benefit the ASR performance.
On the other hand, our model can still predict the intent labels correctly for 83.4\% of the utterances containing ASR errors, showing the robustness of our approach against ASR errors. 

\begin{table}
\centering
\caption{\label{ablation} Ablation study results on SLURP.}
\begin{tabular}{ lc } 
 \hline
 \hline
 Approach & Accuracy\\ 
 \hline
 HuBERT + BiLSTM NLU & 83.95 \\
 HuBERT + BERT NLU & 87.44 \\
 HuBERT + linear classifier w/o ASR loss  &  84.81 \\ 
 HuBERT (frozen) +  utterance encoder &  72.16 \\ 
 HuBERT with probability as input  &  87.00 \\ 
 HuBERT with CNN textual encoder  &  88.05 \\ 
 Ours (HuBERT with logits as input) & 88.18 \\
 \hline 
 \hline 
\end{tabular}
\label{table:ablation}
\end{table}

\section{Conclusion}
In this work, we investigate the use of CTC-based ASR models for utterance level SLU tasks. Experimental results show that joint training with CTC and SLU losses achieves SOTA results on several datasets.
With pretrained acoustic encoders, a small fully connected layer-based utterance encoder can achieve very good performance.
Our approach is also non-auto-regressive, thus efficient at inference time.
For future work, we will investigate how to extend this framework for sequence labeling tasks, like slot filling. The key question is how to recover the entity names from the output of CTC-based ASR models in a differentiable and efficient manner without auto-regressive decoding.

\section{References}
\bibliographystyle{IEEEtran}
{\footnotesize
\bibliography{mybib, refs}}

\end{document}